# Recurrent neural network approach for cyclic job shop scheduling problem

M-Tahar Kechadi a, Kok Seng Low a, G.Goncalves b

a School of Computer Science and Informatics, University College Dublin,
Belfield, Dublin4, Ireland
b University of Artois, LGI2A,F-62400 Behune, France

**Abstract**

While cyclic scheduling is involved in numerous real-world applications, solving the derived problem is still of exponential complexity. This paper focuses specifically on modelling the manufacturing application as a cyclic job shop problem and we have developed an efficient neural network approach to minimise the cycle time of a schedule. Our approach introduces an interesting model for a manufacturing production, and it is also very efficient, adaptive and flexible enough to work with other techniques. Experimental results validated the approach and confirmed our hypotheses about the system model and the efficiency of neural networks for such a class of problems.

## 1. Introduction

Research into scheduling of tasks has spanned from theoretical approaches to their application to real-world manufacturing floors, service sectors and various computing applications. The scheduling problem has been an active research area since 1950s [18]. As part of the combinatorial optimisation problem, there is much interest in this area. The scheduling problem exists not only in particular types of manufacturing environments, but also in the service and management sectors, areas of computing, etc. Moreover, from Conway et al. [19], scheduling problems are proven to be NP-hard. The development of a schedule involves selecting a sequence that guarantees that all required tasks are processed without delays or conflicts. It also involves assigning the required resources needed and the appropriate times to start and complete the processing of each individual task. A good, feasible schedule can impact very much on production costs such as variable production and overtime costs, inventory holding costs, penalty costs associated with missing deadlines, and possible expediting costs for implementing the schedule in a dynamic environment. An efficient scheduling approach may result in high resource utilisation. This is especially true for many production systems with resources of limited capacity, and every fraction of time used contributes to the production system's efficiency. The quantity of work in progress requiring storage space of facilities could potentially be kept to a minimum or even totally eliminated with a good schedule. The chances of lateness can be reduced or eliminated too, as proper scheduling will accurately estimate completion time for all the jobs. This will guarantee timely delivery of orders.

This research is motivated by the challenges of cyclic scheduling problems, which are very complex [1]. The manufacturing system here covers mainly the cyclic job shop. The cyclic job shop scheduling problem, as it was shown by Hanen [2] and Brucker and Kampmeyer [3], is an extended version of the job shop scheduling problem (JSSP), where the jobs are executed repeatedly. JSSP involves the scheduling of jobs, which contain a certain number of operations. Its overall objective is to minimise the schedule length to complete all the jobs. In the case of the cyclic version, the objective is to minimise the cycle time, which is more regular than the general problem.

Early research works into cyclic job shop scheduling (CJSS) were conducted by Roundy [4] on scheduling problems with a limited number of machines that can execute the same job repeatedly. Their goal is to minimise the cycle time of a schedule. Draper et al. [5] utilised constraints satisfaction to find minimum cycle time and work in progress in the CJSS problem. Hanen in [2] also developed a branch and bound approach to deal with a generalised version of the cyclic job shop, by incorporating the precedence constraints, but in the context of computer pipeline. A generalised version of the cyclic job shop with job repetition was studied by Kampmeyer [6], where the height or distance between two adjacent tasks (see Fig. 4); $H_{sink;source} \geq 1$.[1] In these problems, there is more than one occurrence of an operation per cycle time utilising the minimal part set (MPS) Hitz [7]. Another extension to the cyclic job shop problem is the *cyclic job shop problem with blocking* from Brucker and Kampmeyer [3].

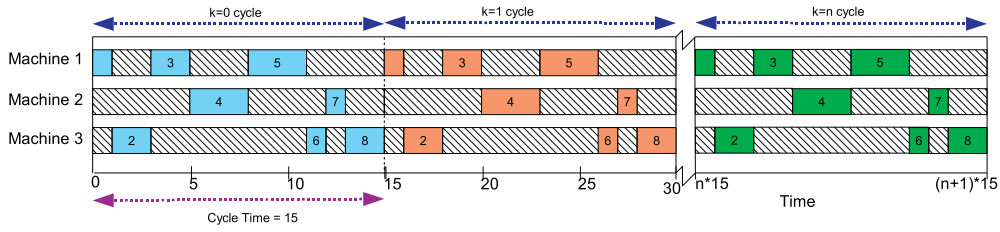

**Fig. 1.** Cyclic schedule of latency equals 1.

This paper is organised as follows: In the next section we discuss the cyclic scheduling problem and its challenges, and define and model precisely the cyclic problem that is studied in this paper. Section 3 introduces our approach to the cyclic scheduling problem that consists not only of a recurrent neural network, but also all the steps required to select a feasible initial schedule, to escape from local minima, and to guarantee the feasibility of the schedules returned by the neural network. In Section 4 we discuss some experimental results based on some benchmarks used in the field. We conclude and give some future directions in Section 5.

## 2. Cyclic scheduling problem

The cyclic scheduling, as a special case of the scheduling problem, can be defined as an infinite number of occurrences of some tasks required to produce numerous pieces of the same type. An occurrence is a single execution of a task. A set of tasks are executed repeatedly over a probable infinite time horizon. Even though the number of task occurrences is infinite, there is a particular cyclic pattern associated with these occurrences. This can be achieved through generating a special framework (or schedule) comprising a pattern of operating sequences, which will be executed repeatedly. This reduces the complexity by seeking out the optimisation of a single schedule. In real-world, many systems, such as production line, have cyclic behaviour.

The cycle time is the time taken to execute all the operations before the next schedule pattern starts. In a cyclic schedule, the difference between two consecutive occurrences of a task is measured as cycle time. Minimising the cycle time is one of the major goals in such systems as it increases the number of jobs completed and maximises the throughput of the machines. The latency is the number of cycles required to complete a single job. This definition also implies the number of jobs concurrently under production. As an example, Fig. 1 shows a cyclic schedule of a job that consists of 8 tasks. The cycle time in this figure has 15 units of time. Basically, as the tasks are linked with precedence constraints (a task $t_i$ of the same occurrence cannot start before the end of the task $t_{i-1}$), the total time needed to execute the 8 tasks is 15 units of time. Fig. 2 shows an improvement of the cycle time for the same system, which is 9 units of time. In order to improve the cycle time we overlap the occurrences of the job. Rather than waiting until the end of an occurrence to start a new one, we can start an occurrence of a job as soon as the machines become available. For instance, the machine 1 in Fig. 2 executes Task 1 of occurrence 3, then Task 3 of occurrence 2 and finally Task 5 of occurrence 1. Of course, we assume, in this execution, that the occurrences 1 and 2 have started before and their corresponding Tasks 1, 2 (for occurrence 2) and Tasks 1, 2, 3, and 4 (for occurrence 1) have been executed in previous cycles. One can notice that the latency in the first case is 1, while it is equal to 4 in the second case. In the second case, in terms of the throughput, we have one job finished every 9 units of time, however, for a specific occurrence of a job, we need 4 cycles to finish its execution.

The cyclic scheduling problem can be divided into 1-periodic schedule (or periodic schedule) and K-periodic schedule for $K > 1$. 1-Periodic schedule has only one occurrence of each operation per cycle and K-periodic schedule has K occurrences of an operation per cycle. An example of a K-periodic schedule is shown in Fig. 3, with $k = 2$. Before going into more details one needs to state the problem constraints and assumptions.

### 2.1. Problem description and assumptions

A cyclic job shop system is generally characterised by the following:

- A set of $M$ machines, $\mathcal{M} = \{m_1, m_2, \ldots, m_M\}$.
- A set of $N$ jobs, $T = \{J_1, J_2, \ldots, J_N\}$.
- Each job $j$ has $N_j$ cyclic operations, $O_j = \{(j, 1), (j, 2), \ldots, (j, N_j)\}$, with their corresponding predefined processing times $\{p_{j1}, p_{j2}, \ldots, p_{jN_j}\}$.
- Each machine has a set of operations to execute and can execute only one operation at a time.
- The operations are linked by precedence constraints.
- Each job has its own unique route through the machines, independently of any other jobs.
- Each operation is assigned to one particular dedicated machine and executed with no interruption and without preemption.

An example of a cyclic job shop that consists of 3 jobs is shown in Fig. 4. Job 1 has 3 operations; {(1, 1), (1, 2), (1, 3)}, Job 2 has 2 operations; {(2, 1), (2, 2)} and Job 3 also has 2 operations; {(3, 1), (3, 2)}. The processing time of each operation is known prior to its execution. In this example, the processing times of the operations are $p_{11} = 6$, $p_{12} = 5$, $p_{13} = 3$, $p_{21} = 1$, $p_{22} = 2$, $p_{31} = 5$ and $p_{32} = 3$. There are 3 machines; $M_1, M_2, M_3$ and the operations are assigned to the machines as follows: the operations (1, 1) and (3, 2) are processed on $M_1$, (1, 2) and (2, 1) are on $M_2$, and the operations (1, 3), (2, 2), and (3, 1) are on $M_3$. The operations are constrained by certain precedence relations; (1, 1) precedes (1, 2), operation (1, 2) precedes (1, 3), operation (2, 1) precedes (2, 2) and operation (3, 1) precedes (3, 2).

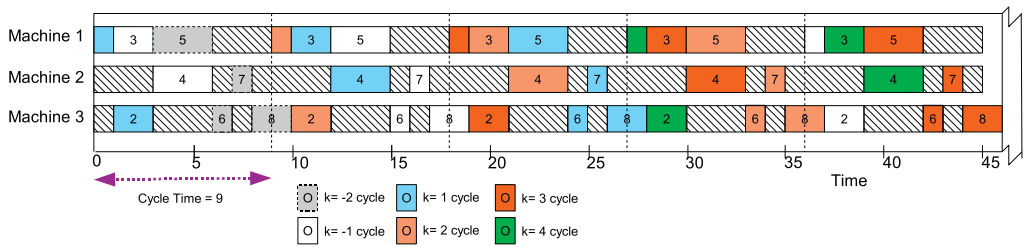

**Fig. 2.** Cyclic schedule of latency equals 4.

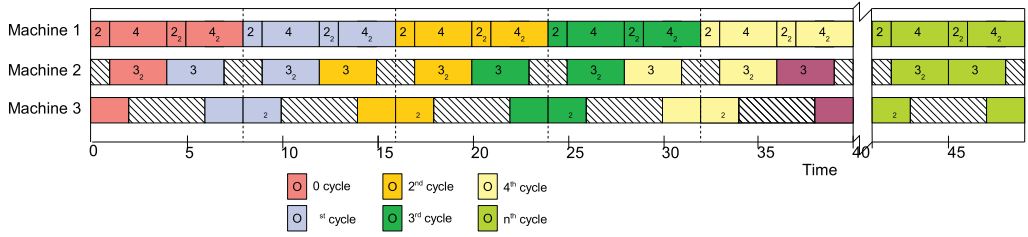

**Fig. 3.** 2-Periodic cyclic schedule.

Our approach is based on the following assumptions or conditions of the application:

1. The information relating to the jobs and machines is available.
2. The processing time required by the operations are known.
3. All jobs and operations are available at start time.
4. Unlimited storage space is available, so no penalties associated with jobs waiting.
5. The job operations are processed in a predetermined order.
6. All set-up times on the machines are included in the processing time.

While the assumption 1 is obvious, the second one is not straightforward, as the processing times are not easy to determine in real-world applications. But they can always be estimated using an efficient pre-processing technique. The remaining assumptions attempt to exclude some special cases, such as stochastic effects, and random execution of the operations. These assumptions allow an accurate modelling and representation of the cyclic job shop scheduling (CJSS) problem.

In the rest of the paper we will focus on periodic schedules. The period of a schedule is equal to the cycle time. In the following we will describe a cyclic job shop scheduling and highlight its characteristics that can help in developing a model for it.

### 2.2. CJSS Model

Let $S_{ij}^k$ be the start time of the occurrence $k$ of the operation $(i, j)$ of a job $i$ and let $p_{ij}$ be its processing time.

**Definition 1.** Let $\tau$ be the cycle time. A cyclic schedule is periodic if the following holds true for any start time of an operation $(i, j)$ of a job $i$:

$$S_{ij}^k = S_{ij}^0 + \tau k; \quad \forall k \in \mathbb{N} \tag{1}$$

**Definition 2.** The *disjunctive constraints* for every machine are defined between every pair of its operations $(i, j)$ and $(e, f)$ as follows:

$$(S_{ij}^k + p_{ij}) \leq S_{ef}^h \quad \text{or} \quad (S_{ef}^h + p_{ef}) \leq S_{ij}^k \quad k, h \in \mathbb{N} \tag{2}$$

One can notice from this definition is that if the operation $(i, j)$ is processed before $(e, f)$ then $S_{ij}^k + p_{ij} \leq S_{ef}^h$ is true, otherwise $(e, f)$ is processed before $(i, j)$, so $S_{ef}^h + p_{ef} \leq S_{ij}^k$ must be true. Due to the cyclic nature of this system, one has to consider also the disjunctive constraints in relation to the occurrences of the same operation. This case is also reflected in the above definition, where $i = e$ and $j = f$.

To illustrate the disjunctive constraints, we consider the example given in Fig. 4. For instance, the machine $m_3$ has three operations

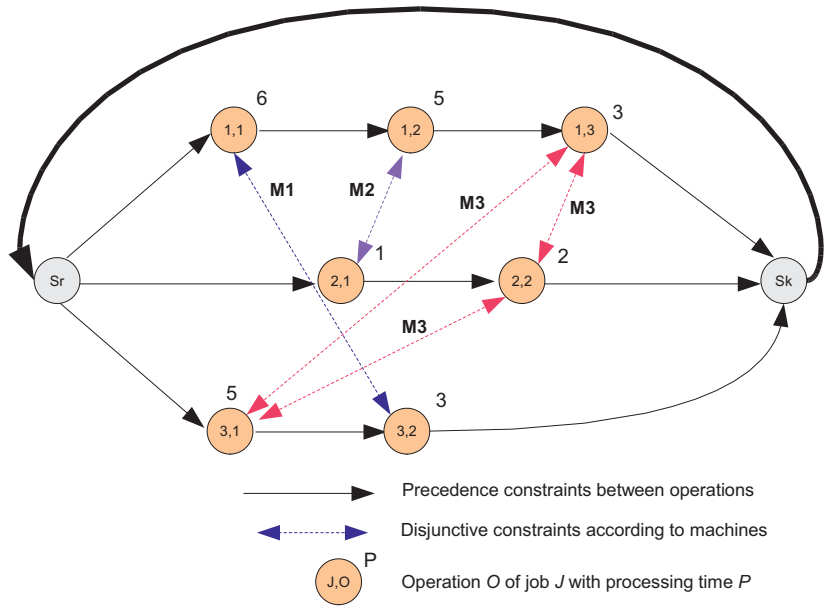

**Fig. 4.** Example of a cyclic job shop, with 3 jobs; {(1, 1), (1, 2), (1, 3)}, {(2, 1), (2, 2)}, and {(3, 1), (3, 2)}.

(1, 3), (2, 2), and (3, 1). The disjunctive constraints on $M_3$ for these operations are as follows:

$$\delta_{3113}(S_{31}^k - S_{13}^l + p_{31}) \leq 0$$
$$(1 - \delta_{3113})(S_{13}^l - S_{31}^k + p_{13}) \leq 0$$
$$\delta_{3122}(S_{31}^k - S_{22}^l + p_{31}) \leq 0$$
$$(1 - \delta_{3122})(S_{22}^l - S_{31}^k + p_{22}) \leq 0$$
$$\delta_{1322}(S_{13}^k - S_{22}^l + p_{13}) \leq 0$$
$$(1 - \delta_{1322})(S_{22}^l - S_{13}^k + p_{22}) \leq 0$$

where $\delta_{ijef}$ is the Kronecker symbol defined by

$$\delta_{ijab} = \begin{cases} 1 & \text{if } S_{ij} - S_{ef} \leq 0 \\ 0 & \text{Otherwise} \end{cases} \quad (3)$$

**Definition 3.** The *precedence constraints* (or conjunctive constraints) are defined between operations of the same job. If the operation $(i, j)$ should be executed before $(i, l)$, then we can write:

$$S_{ij}^k - S_{il}^k + p_{ij} \leq 0 \quad (4)$$

For the example given in Fig. 4, the job 1 has the following precedence constraints:

$$S_{11} - S_{12} + p_{11} \leq 0$$
$$S_{12} - S_{13} + p_{12} \leq 0 \quad (5)$$

The end of a particular schedule is taken as the point of the completion time of the last operation. In other words, this corresponds to the maximum completion time of the jobs. The start and completion times of a particular job $i$ can be calculated as follows:

$$S_i = \min_{j=1}^{N_i}(S_{ij}) \quad \text{and} \quad C_i = \max_{j=1}^{N_i}(S_{ij} + p_{ij}) \quad (6)$$

The start time and the maximum completion time of all the jobs are:

$$S_{min} = \min_{i=1}^{N}\{S_i\} \quad \text{and} \quad C_{max} = \max_{i=1}^{N}\{C_i\} \quad (7)$$

where $N$ is the number of jobs in the system. The cycle time can then be calculated as the difference between the maximum completion time and the earliest start time of the jobs: $\tau = C_{max} - S_{min}$. This equation does not guarantee the minimum cycle schedule. In order to do so, we need to model the system as a constrained optimisation problem. Let $S$ be the schedule that consists of start times of the operations of all jobs; a vector of stat times: $S = (S_{11}, \ldots, S_{1N_1}, \ldots, S_{N1}, \ldots, S_{NN_k})$. The system can be formulated as follows:

$$\text{Minimise } F(S) = \sum_{i=1}^{N}\sum_{j=1}^{N_i}(S_{ij}^{k+1} - S_{ij}^k) \quad (8)$$

Subject to

$$S_{ij}^k - S_{il}^k + p_{ij} \leq 0 \quad (i, j), (i, l) \in O_i$$
$$S_{ij}^k \geq 0 \quad i, j, k \geq 0$$
$$\delta_{ijab}(S_{ij}^k - S_{ab}^l + p_{ij}) \leq 0 \quad ((i, j), (a, b)) \in D_m, \ m=1,\ldots,M$$
$$(1 - \delta_{ijab})(S_{ab}^l - S_{ij}^k + p_{ab}) \leq 0 \quad ((i, j), (a, b)) \in D_m, \ m=1,\ldots,M$$

where $O_i$ is the set of operations of the job $i$ and $D_m$ a set of operations, which are assigned to the machine $m$.

## 3. Recurrent neural network model for CJSS

The structure of a neural network (NN) consists of largely interconnected processing units (or neurons), usually organised as layers. The connections between neurons of each layer, and between layers, the number of layers and neurons in each layers, as well as the learning technique used to train it give to that neural network its characteristics, ability, and flexibility to model real-world systems and even define the different classes of neural networks. For instance, a recurrent neural network (RNN) is a type of a feedforward network in which the neurons recurrently working on internal states. In other words, it uses feedback that includes the initial and past states to derive the next state. Common recurrent neural networks are the Hopfield networks, competitive networks, Self-Organising-Maps (SOM) and Constraints Satisfaction Adaptive Neural Networks.

Several researchers [8–10] have applied neural networks on scheduling problems. Many of these works focused on the types of neural network to use either for production or manufacturing scheduling problems. However, there is very little done in the case of the cyclic scheduling. In this paper, we propose to use RNN for CJSS, as it has two very interesting features for this problem:

- RNNs are type of feedforward and feedback neural networks, which have the ability to better model the non-linearity embedded in most real-world problems.
- RNN is well adapted for unsupervised learning,

The RNN model [11] greatly depends on the mathematical model used. In order to use a neural network technique, the optimisation problem defined above (8) should be transformed to its equivalent unconstrained optimisation problem. This transformation is accomplished through the inclusion of the constraints in the objective function, called penalty function.

The inequality constraints described in the problem (8) can be rewritten as

$$r_l(S) = S_{ij}^k - S_{ih}^k + p_{ij} \quad \forall k \geq 0 \quad \text{and} \quad (i, j), (i, h) \in O_i \quad (9)$$

$$r_{\gamma+l}(S) = S_{ij}^k - S_{ab}^h + p_{ij} \quad \forall k, h \geq 0 \quad \text{and} \quad (i, j), (a, b) \in \mathcal{D} \quad (10)$$

where $\gamma$ is the total number of conjunctive constraints and its upper bound is by $\sum_{i=1}^{N}N_i - N$. $\omega$ is the total number of disjunctive constraints in $\mathcal{D}$ and it is equal to $(1/2)(\sum_{i=1}^{M}m_i^2 - M)$. $M$ is the number of machines in the system, and $m_i$ is the number of operations allocated to the machine $i$. We denote $\xi = \gamma + \omega$.

The energy function needs to be constructed so that it penalises every violation of the constraints. We Consider the energy function of the form

$$E(S) = F(S) + KP(S) \quad (11)$$

where $F(S)$ is the cost function given in Eq. (8), $K$ is the penalty factor and $P(S)$ is the penalty function of the constraints, which can be expressed as $\sum_{i=1}^{\xi}\varphi[r_i(S)]$. So, $E(S)$ can be written as follows:

$$E(S) = \sum_{j=1}^{N}\sum_{i=1}^{N_j}(S_{ij}^{k+1} - S_{ij}^k) + K\sum_{i=1}^{\xi}\varphi[r_i(S)] \quad (12)$$

where

$$\varphi[r_i(S)] \begin{cases} = 0 & \text{if } r_i(S) \leq 0 \\ > 0 & \text{if } r_i(S) > 0 \end{cases} \quad (13)$$

The function $\varphi(\vartheta)$ has been chosen so that the property (13) is fulfilled. $K$ is a positive parameter that controls the scaling factor between the cost term and the penalty term of the unconstrained optimisation problem in (11). The scaling factor should be chosen such that the optimal solution can be reached while the constraint violation is penalised. It can be easily shown that the two problems become equivalent as ($K \to +\infty$).

### 3.1. The proposed RNN

Initially, RNN takes as input the vector of start times; the schedule. As RNN is recursive, the decision variables (i.e., start times of the operations) are recursively updated and/or improved. The calculations of the decision variables are streamed through the network, where some changes are applied to the previous solutions. The penalty function in the network that encompasses the constraints will dictate the direction of changes for the decision variable values.

RNN models the energy function deduced from the cyclic job shop scheduling problem. This function is given in Eq. (12). We use Lagrangian Relaxation method to build our RNN, based on Eq. (11). We also choose a quadratic form $((1/2)x^2)$ for the function $\varphi(.)$ in order to fulfil the constraint inequality (13). By applying the steepest descent approach, we can generate the updates of the schedule by calculating the following equation:

$$S(t+1) = S(t) - \mu \frac{\partial E(S)}{\partial S} \qquad (14)$$

where $S(t)$ is the schedule at time or iteration $t$ and $\mu$ is the learning rate. Eq. (14) is called the equation of motion and describes the dynamics of the network.

#### 3.1.1. Schedule perturbation phase

In order to improve the solutions found as well as the capability of RNN described above (see Eq. (14), we incorporate a *schedule perturbation phase* into the network. The proposed perturbation phase is similar to the perturbation used in the Simulated Annealing approach. However in our approach, the perturbation phase is only activated based on a special condition, which is when the system is identified to be trapped in a local minimum. A local minimum state is defined as a state in which the system is stable but it may not necessarily be the global minimum state. This schedule perturbation works as follows:

- **Phase 1:** Identify the local minimum state reached. This is done by comparing the energy values for the last few iterations. In other words, identify situations in which the system is not evolving at all.
- **Phase 2:** Generate a corresponding perturbed factor.
- **Phase 3:** Unsettle the current state of the solution by incorporating the perturbed factor into the current solution.
- **Phase 4:** Release new generated solution to the network and let it find another stable state.

The schedule perturbation phase is shown in Algorithm 1.

**Algorithm 1.** Perturbation schedule algorithm.

---
**Input**: List of Schedule $S$ at iteration $n$, Set $E^{(n-2)}, E^{(n-1)}, E^{(n)}$, parameter $\pi$, parameter $Threshold$
**Output**: Schedule with perturbed start times
**if** $(E^{(n-1)} - E^{(n-2)} < Threshold) \wedge (E^{(n)} - E^{(n-1)} < Threshold)$ **then**
    **foreach** *operation o in List S* **do**
        Apply processing time, $p^*_{i;j} = p_{i;j} + \Gamma(-\pi p_{i;j}, \pi p_{i;j})$ ;
        Apply start times, $s^*_{i,j} = s_{i;j} + p^*_{i;j}(G(1-\Omega))$ ;
        Update start time to List $S$ ;
    **end**
**end**

---

#### 3.1.2. Competitive dispatch rule phase

Although RNN has been augmented with a schedule perturbation, we also preprocess the initial schedule before feeding it to the network. This preprocessing phase is called **competitive dispatch rule phase (CDRP)** and its main goal is to generate an initial schedule that is feasible. The preprocessing procedure is a set of rules that can be applied on the initial schedule. The rules are selected based on the properties associated with the jobs or operations. We have used two rules: **weighted shortest processing time (WSPT)** and **weighted longest processing time (WLPT)**. They are described below.

- **Weighted shortest processing time (WSPT)**: This rule is described the Algorithm 2. The WSPT rule is one of the simplest and most effective in the system design. This rule is commonly used to minimise the total completion time, mean flow time and percentage of tardy jobs citeconway67. The WSPT rule has been identified in the literature to perform very well in terms of minimising the weighted mean flow time as well as reducing the percentage of tardy jobs, especially under highly loaded conditions. Comparative studies have listed WSPT as one of the most consistent rules in solving job shop problems.

**Algorithm 2.** Weighted shortest processing time algorithm.

---
**Input**: List of Schedule $S$
**Output**: Schedule with WSPT Algorithm
Initialise $StartTime\ S := 0$ ;
Initialise $CompletionTime_{job=j} := 0$;
**foreach** *job j from List S* **do**
    **foreach** *operation o in job j* **do**
        Find the operation $o$ with smallest processing time, $p_{i;j}$ ;
        Place in assigned machine $M_{m=o}$ from List $M$ ;
        $StartTime = \max\{CompletionTime_{job=j}, CompletionTime_{M_{m=o}}\}$;
        Update $CompletionTime_{job=j} = StarTime + p_{i;j}$;
        Update $CompletionTime_{M_{m=o}} = StarTime + p_{i;j}$;
        Assign operation to beginning of list $S$;
    **end**
**end**

---

- **Weighted longest processing time (WLPT):** As opposed to the WSPT rule, this rule will load the job operations with the largest processing onto their respective machines.

CDRP works through a parallel single run of each dispatch rule. This will generate three feasible schedules. Then based on the objective function the best among the three schedules will be selected as the initial solution.

### 3.1.3. Schedule post-processing phase

This phase is introduced to complement the solutions returned by the neural network model. Due to some limitations associated with the quality of solutions (the network does not guarantee feasibility), it is important that the produced schedules are all feasible. In addition to guaranteeing feasible solution $S$, this procedure also aims to achieve the following:

1. Check and eliminate any minor violations associated with resource (or disjunctive) constraints. This is completed using the **adhere disjunctive algorithm** given in Algorithm 3.
2. Check and eliminate any minor violations associated with precedence (or conjunctive) constraints. This is achieved through the **adhere conjunctive algorithm**, which is similar to Algorithm 3.
3. Transform any optimal solutions obtained by RNN to the earliest start time state using **compact schedule algorithm**.

**Algorithm 3.** Adhere disjunctive algorithm.

```
Input: Start Times
Output: Schedule adhering to Disjunctive Constraints
for Machine m:=1 to M do
    for OperationOnMachine_m O(m) = 1 to N(m − 1) do
        if StartTime_{O(m)+1} − (StartTime_{O(m)} + p_{O(m)}) < 0 then
            Δ:=StartTime_{O(m)+1} − (StartTime_O(m) + p_O(m)) ;
            for OperationOnMachine_m O(m) = 1 to N(m) do
                StartTime_O(m + 1):=StartTime_O(m + 1) + Δ;
            end
        end
    end
end
```

The **Post-processing** step is vital to ensure that the stat-time of operations are discrete values.

## 4. Experimental results

Our techniques were implemented in JAVA and the evaluation is carried out on some very popular benchmarks. These benchmarks are described below. Depending on the benchmark characteristics the evaluation is measured based on the following criteria:

1. The accuracy of solutions found against known optimal results. This is measured by percentage deviation of solution, which is given by

$$\rho = \frac{(\text{Best Solution Found}) - (\text{Optimal Solution})}{(\text{Optimal Solution})} \times 100 \quad (15)$$

2. Percentage improvements in terms of response time and the number of iterations are needed to reach optimal solution.
3. The response time for obtaining optimal solution.

### 4.1. Benchmarks

Several benchmark problems were used to evaluate our approach and compare it to the existing techniques. The selected benchmark problems cover various aspects of the evaluation and validation. These benchmarks were initially designed for the general job shop scheduling techniques, and they were modified to make them cyclic without loss of any of their characteristics. The reasons for this are twofold: (1) there is a lack of benchmarks designed specifically for the cyclic scheduling techniques, and (2) the selected benchmarks were proved to be very important in the evaluation of the job shop scheduling tasks in general.

The benchmarks we used have already been published by Fisher and Thompson [12], Lawrence [13], Adams et al. [14], Applegate and Cook [15] and Storer et al. [16] and they are summarised in the following:

FT06, FT10, FT20: These are used by Fisher and Thompson [12]. FT06, FT10, and FT20 consist of 6 Jobs and 6 Machines, 10 Jobs and 10 Machines, and 20 Jobs and 5 Machines, respectively. The processing times for the operations for all these benchmarks are in the range of [1 : 99]. Table 1 shows more details about these problems.

LA01–LA40: There are 40 variants of problems proposed by Lawrence in [13]. They are divided into 8 groups of 5 problems each. The first group consists of 10 Jobs and 5 Machines (10 × 5). The second, third, until eighth groups are of sizes (15 × 5), (20 × 5), (10 × 10), (15 × 10), (20 × 10), (30 × 10), and (15 × 15), respectively. The processing times for the operations are in the range of [5 : 99]. For more details, the reader can find these problems in [12].

ABZ5–ABZ9: Adams et al. [14] generated 5 problem instances. Two of these instances consist of 10 Jobs and 10 Machines, the remaining three are of size (20 × 15). The processing times for the operations are in the range of [11 : 100].

ORB01–ORB10: Applegate and Cook [15] proposed 10 problem instances of 10 Jobs and 10 Machines each. The processing times for the operations are in the range of [1 : 99].

**Table 1**
Benchmark CJSS problem from [12] (CC: conjunctive constraints, DC: disjunctive constraints).

| Problem | No. jobs (N) | No. M (M) | No. Ops | No. CC | No. DC |
| --- | --- | --- | --- | --- | --- |
| FT06 | 6 | 6 | 36 | 30 | 90 |
| FT10 | 10 | 10 | 100 | 90 | 450 |
| FT20 | 20 | 5 | 100 | 80 | 950 |

While the sizes of all these problems are of the same order of magnitude, they differ hugely in terms of the number of operations in each job, the number of conjunctive and disjunctive constraints. We have used all these benchmarks in our testing phase. The experiments were executed with several termination criteria and no limit in terms of the execution time required.

### 4.2. CJSS results

In this section, we present the results of our approach on various benchmarks introduced above. We also show the optimal solutions found for each instance of a given benchmark problem. Furthermore we will also analyse the performance of both approaches. We will compare our solutions with the best results reported thus far by other researchers.

Table 2 shows the best results found for the benchmark problems LA01–LA40. Table 3 shows the best results obtained for the problems FT06, FT10, and FT20. Table 4 shows results obtained for problems ABZ5–ABZ9. Finally Table 5 shows results obtained for the 10 benchmarks ORB01–ORB10. We have reported the best results in blue colour when the best solution is equal to the known computed lower bound, which means that we have found the optimal solution.

As we can see, both RNN and LRRNN approaches returned optimal solutions for the problems FT06, FT10, FT20 and LA01–LA10. Moreover, the LRRNN approach reported the best results for problems LA11–LA15, and performed competitively well for large problems containing more than 100 operations as shown in LA16–LA28 (100–200 operations). For the problems ORB01–ORB10, LRRNN outperforms RNN in finding optimal solutions for 5 out of the 10 problems (ORB02, ORB03, ORB04, ORB05 and ORB07). However both RNN and LRRNN approaches under-performed in the case of the 5 ABZ problems (ABZ5–ABZ9).

**Table 2**
Best results found for Lawrence benchmarks.

| Prob. | $CT_{LB}$ | RNN | | | | LRNN | | | | Δ |
|---|---|---|---|---|---|---|---|---|---|---|
| | | $CT_{Best}$ | $It_{Best}$ | T (s) | MRE | $CT_{Best}$ | $It_{Best}$ | T (s) | MRE | |
| *10 Jobs × 5 Machines, 50 Operations* | | | | | | | | | | |
| LA01 | 666 | 666 | 4947 | 9.12 | 0.0% | 666 | 5023 | 9.26 | 0.0% | 0.0% |
| LA02 | 655 | 655 | 11,896 | 21.93 | 0.0% | 655 | 12,043 | 22.20 | 0.0% | 0.0% |
| LA03 | 597 | 597 | 133,883 | 246.80 | 0.0% | 597 | 135,409 | 249.62 | 0.0% | 0.0% |
| LA04 | 590 | 590 | 130,742 | 241.01 | 0.0% | 590 | 132,337 | 243.95 | 0.0% | 0.0% |
| LA05 | 593 | 593 | 2042 | 3.76 | 0.0% | 593 | 2078 | 3.83 | 0.0% | 0.0% |
| *15 Jobs × 5 Machines, 75 Operations* | | | | | | | | | | |
| LA06 | 926 | 926 | 5026 | 9.26 | 0.0% | 926 | 5129 | 9.45 | 0.0% | 0.0% |
| LA07 | 890 | 890 | 5929 | 10.93 | 0.0% | 890 | 6047 | 11.15 | 0.0% | 0.0% |
| LA08 | 863 | 863 | 17,746 | 32.71 | 0.0% | 863 | 17,931 | 33.05 | 0.0% | 0.0% |
| LA09 | 951 | 951 | 3337 | 6.15 | 0.0% | 951 | 3418 | 6.30 | 0.0% | 0.0% |
| LA10 | 958 | 958 | 54,967 | 101.33 | 0.0% | 958 | 56,126 | 103.46 | 0.0% | 0.0% |
| *20 Jobs × 5 Machines, 100 Operations* | | | | | | | | | | |
| LA11 | 1222 | 1562 | 7970 | 14.69 | 27.8% | 1222 | 8334 | 15.36 | 0.0% | 21.8% |
| LA12 | 1039 | 1039 | 3416 | 6.30 | 0.0% | 1039 | 3542 | 6.53 | 0.0% | 0.0% |
| LA13 | 1150 | 1150 | 4829 | 8.90 | 0.0% | 1150 | 4955 | 9.13 | 0.0% | 0.0% |
| LA14 | 1292 | 1487 | 3691 | 6.80 | 15.1% | 1292 | 3778 | 6.96 | 0.0% | 13.1% |
| LA15 | 1207 | 1402 | 12,132 | 22.36 | 16.2% | 1207 | 12,735 | 23.48 | 0.0% | 13.9% |
| *10 Jobs × 10 Machines, 100 Operations* | | | | | | | | | | |
| LA16 | 945 | 982 | 231,645 | 427.02 | 3.9% | 957 | 235,050 | 433.30 | 1.3% | 2.5% |
| LA17 | 784 | 784 | 369,062 | 680.34 | 0.0% | 784 | 375,963 | 693.06 | 0.0% | 0.0% |
| LA18 | 848 | 848 | 416,176 | 767.19 | 0.0% | 848 | 422,169 | 778.24 | 0.0% | 0.0% |
| LA19 | 842 | 859 | 451,512 | 832.33 | 2.0% | 842 | 459,368 | 846.81 | 0.0% | 2.0% |
| LA20 | 902 | 1052 | 677,898 | 1249.65 | 16.6% | 908 | 690,032 | 1272.02 | 0.7% | 13.7% |
| *15 Jobs × 10 Machines, 150 Operations* | | | | | | | | | | |
| LA21 | 1046 | 1248 | 277,687 | 511.89 | 19.3% | 1074 | 289,905 | 534.42 | 2.7% | 13.9% |
| LA22 | 927 | 1047 | 718,493 | 1324.48 | 12.9% | 932 | 748,957 | 1380.64 | 0.5% | 11.0% |
| LA23 | 1032 | 1059 | 883,393 | 1628.46 | 2.6% | 1054 | 925,707 | 1706.47 | 2.1% | 0.5% |
| LA24 | 935 | 1024 | 877,567 | 1617.72 | 9.5% | 944 | 920,392 | 1696.67 | 1.0% | 7.8% |
| LA25 | 977 | 1049 | 385,552 | 710.73 | 7.4% | 984 | 405,523 | 747.55 | 0.7% | 6.2% |
| *20 Jobs × 10 Machines, 200 Operations* | | | | | | | | | | |
| LA26 | 1218 | 1322 | 208,088 | 383.59 | 8.5% | 1224 | 259,669 | 478.68 | 0.5% | 7.4% |
| LA27 | 1235 | 1276 | 766,889 | 1413.70 | 3.3% | 1249 | 1,168,739 | 2154.48 | 1.1% | 2.1% |
| LA28 | 1216 | 1287 | 1,197,488 | 2207.47 | 5.8% | 1235 | 1,506,440 | 2777.00 | 1.6% | 4.0% |
| LA29 | 1152 | 1458 | 1,377,908 | 2540.06 | 26.6% | 1249 | 1,824,350 | 3363.04 | 8.4% | 14.3% |
| LA30 | 1355 | 1396 | 776,778 | 1431.93 | 3.0% | 1355 | 1,022,706 | 1885.28 | 0.0% | 2.9% |
| *30 Jobs × 10 Machines, 300 Operations* | | | | | | | | | | |
| LA31 | 1784 | 2189 | 150,373 | 277.20 | 22.7% | 1855 | 162,613 | 299.76 | 4.0% | 15.3% |
| LA32 | 1850 | 2478 | 114,252 | 210.61 | 33.9% | 1947 | 123,312 | 227.32 | 5.2% | 21.4% |
| LA33 | 1719 | 2416 | 100,510 | 185.28 | 40.5% | 1840 | 111,707 | 205.92 | 7.0% | 23.8% |
| LA34 | 1721 | 2441 | 108,363 | 199.76 | 41.8% | 1849 | 109,880 | 202.55 | 7.4% | 24.3% |
| LA35 | 1888 | 2511 | 83,628 | 154.16 | 33.0% | 1994 | 88,930 | 163.94 | 5.6% | 20.6% |
| *15 Jobs × 15 Machines, 225 Operations* | | | | | | | | | | |
| LA36 | 1268 | 1468 | 1,911,586 | 3523.86 | 15.8% | 1299 | 2,455,623 | 4526.75 | 2.4% | 11.5% |
| LA37 | 1397 | 1574 | 1,619,023 | 2984.54 | 12.7% | 1405 | 2,225,509 | 4102.55 | 0.6% | 10.7% |
| LA38 | 1196 | 1369 | 1,647,812 | 3037.61 | 14.5% | 1257 | 2,238,717 | 4126.90 | 5.1% | 8.2% |
| LA39 | 1233 | 1548 | 1,264,812 | 2331.58 | 25.5% | 1302 | 1,843,413 | 3398.18 | 5.6% | 15.9% |
| LA40 | 1222 | 1325 | 1,589,312 | 2929.77 | 8.4% | 1320 | 1,903,138 | 3508.28 | 8.0% | 0.4% |

**Table 3**
Best results found for benchmarks FT06, FT10, and FT20.

| Prob | $CT_{LB}$ | DC | CC | RNN | | | | LRNN | | | | %Improv |
|---|---|---|---|---|---|---|---|---|---|---|---|---|
| | | | | $CT_{Best}$ | $It_{Best}$ | T (s) | MRE | $CT_{Best}$ | $It_{Best}$ | T (s) | MRE | |
| *FT Problems, 6 Jobs × 6 Machines, 36 Operations* | | | | | | | | | | | | |
| FT06 | 55 | 30 | 90 | **55** | 336 | 0.62 | 0 | **55** | 339 | 0.62 | 0% | 0% |
| *FT Problems, 10 Jobs × 10 M, 100 Operations* | | | | | | | | | | | | |
| FT10 | 930 | 90 | 450 | **930** | 18,815 | 34.62 | 0 | **930** | 19,087 | 35.12 | 0% | 0% |
| *FT Problems, 20 Jobs × 5 M, 100 Operations* | | | | | | | | | | | | |
| FT20 | 1165 | 80 | 950 | **1165** | 8092 | 14.89 | 0 | **1165** | 8210 | 15.10 | 0% | 0% |

**Table 4**
Best results found for benchmark Adams et al. [14].

| Prob | $CT_{LB}$ | RNN | | | | LRNN | | | | Δ |
|---|---|---|---|---|---|---|---|---|---|---|
| | | $CT_{Best}$ | $It_{Best}$ | T (s) | MRE | $CT_{Best}$ | $It_{Best}$ | T (s) | MRE | |
| *ABZ Problems, 10 Jobs × 10 Machines, 100 Operations* | | | | | | | | | | |
| ABZ5 | 1234 | 1455 | 84,898 | 156.50 | 17.9% | 1297 | 91,367 | 168.12 | 5.1% | 10.9% |
| ABZ6 | 943 | 1032 | 675,548 | 1245.32 | 9.4% | 978 | 721,147 | 1326.91 | 3.7% | 5.2% |
| *ABZ Problems, 20 Jobs × 15 Machines, 300 Operations* | | | | | | | | | | |
| ABZ7 | 656 | 748 | 1,229,878 | 2267.18 | 14.0% | 694 | 1,325,808 | 2439.49 | 5.8% | 7.2% |
| ABZ8 | 645 | 745 | 554,231 | 1021.68 | 15.5% | 697 | 585,157 | 1076.69 | 8.1% | 6.4% |
| ABZ9 | 661 | 786 | 588,761 | 1085.33 | 18.9% | 694 | 620,554 | 1141.82 | 5.0% | 11.7% |

Although our RNN approach performed well for cyclic job shop problems with cycle time up to 1165 (i.e., in the case of the FT20 problem) and a total number of constraints up to 585. This particular approach does not scale well and starts under-performing for larger problems with lower bound cycle time above 1165, for instance. This is obvious in the case of problems with 300 operations (i.e., problems LA31–LA35 where the RNN approach produced an average deviation from the optimal results of about 34.4%. However this could have been greatly influenced by the fact that these problems have the largest total number of constraints, among all the problems considered, consisting of 270 conjunctive constraints and 4350 disjunctive constraints.

Table 6 shows the performance analysis of both RNN and LRRNN approaches. In analysing the computational time for the two approaches we can see that in the case of RNN, it requires an average of 104 s in the platform environment in which they were tested. However the RNN approach may end at an early stage of its execution when the problem is too complex (e.g., ends at average of 205 s for the problems LA31–35 with MRE of 34.43%), this is because it was trapped in a local minimum that the perturbation phase is unable to deal with.

On the other hand, in the case of LRRNN, this approach may also suffer from the same setback of retiring early for very complex problems without reaching the optimal solution (e.g., it was stuck at an average of 219 s for the LA31–35 problems with an MRE of 5.86%). Although this could be problem-specific as in the case of nearly similar problems such as ABZ7–ABZ9 (300 operations), LRRNN had taken an average of 1552 s to obtain the best solution with MRE of 6.28%. When we analyse the computational time it takes for both approaches to reach optimal solutions in small problems, the RNN approach is faster than the LRRNN approach. This is obvious in the case of FT06, FT10 and FT20 where RNN was 1.61%, 1.44% and 1.48% faster than LRRNN.

Overall, the LRRNN approach outperforms RNN in nearly all CJSSP problems by an average of 5.14%. Figs. 5–7 show that the deviation from the optimum solutions for all CJSSP benchmarks for both approaches (RNN and LRRNN) (Tables 7 and 8).

We can conclude that the RNN and LRRNN approaches are able to solve cyclic job shop scheduling problems averaging 100 operations. The RNN and LRRNN under-performed for problems with more than 200 operations due to the complexity of constraints involved. Overall, one can notice that RNN is faster in computing optimum solutions for small problems than the LRRNN approach. It does not scale well. On the other hand LRRNN performs much better for larger sizes than RNNA, and it scales much better than RNN (Fig. 8).

**Table 5**
Best results found for benchmark Applegate and Cook [15].

| Prob. | $CT_{LB}$ | RNN | | | | LRNN | | | | Δ |
|---|---|---|---|---|---|---|---|---|---|---|
| | | $CT_{Best}$ | $It_{Best}$ | T (s) | MRE | $CT_{Best}$ | $It_{Best}$ | T (s) | MRE | |
| *ORB Problems, 10 Jobs × 10 Machines, 100 Operations* | | | | | | | | | | |
| ORB01 | 1059 | 1076 | 77,898 | 143.60 | 1.6% | 1061 | 83,834 | 154.25 | 0.2% | 1.4% |
| ORB02 | 888 | 923 | 85,452 | 157.52 | 3.9% | 888 | 86,067 | 158.36 | 0.0% | 3.8% |
| ORB03 | 1005 | 1022 | 94,556 | 174.31 | 1.7% | 1005 | 101,761 | 187.24 | 0.0% | 1.7% |
| ORB04 | 1005 | 1019 | 76,523 | 141.06 | 1.4% | 1005 | 82,354 | 151.53 | 0.0% | 1.4% |
| ORB05 | 887 | 892 | 39,006 | 71.90 | 0.6% | 887 | 41,557 | 76.46 | 0.0% | 0.6% |
| ORB06 | 1010 | 1018 | 77,898 | 143.60 | 0.8% | 1029 | 83,694 | 154.00 | 1.1% | −1.9% |
| ORB07 | 397 | 402 | 66,578 | 122.73 | 1.3% | 397 | 68,708 | 126.42 | 0.0% | 1.2% |
| ORB08 | 899 | 913 | 88,009 | 162.24 | 1.6% | 921 | 95,842 | 176.35 | 2.4% | −0.9% |
| ORB09 | 934 | 956 | 75,447 | 139.08 | 2.4% | 966 | 78,125 | 143.75 | 3.4% | −1.0% |
| ORB10 | 944 | 959 | 66,591 | 122.76 | 1.6% | 969 | 69,035 | 127.02 | 2.6% | −1.0% |

**Table 6**
Performance analysis for CJSSP benchmarks for RNN and LRRNN approaches.

| Problem | N | M | O | Number of ConjConstr | Number of DisjCons | RNN Average CPU(s) | RNN Average $MRE(\%)_{RNN}$ | LRNN Average CPU(s) | LRNN Average $MRE(\%)_{LRRNN}$ | LRNN Average %Improv | LRNN Average %ImprComT |
|---|---|---|---|---|---|---|---|---|---|---|---|
| FT06 | 6 | 6 | 36 | 30 | 90 | 0.62 | 0.00% | 0.63 | 0.00% | 0.00% | −1.61% |
| FT10 | 10 | 10 | 100 | 90 | 450 | 34.62 | 0.00% | 35.12 | 0.00% | 0.00% | −1.44% |
| FT20 | 20 | 5 | 100 | 80 | 950 | 14.89 | 0.00% | 15.11 | 0.00% | 0.00% | −1.48% |
| LA01–LA05 | 10 | 5 | 50 | 40 | 225 | 104.53 | 0.00% | 105.77 | 0.00% | 0.00% | −1.19% |
| LA06–LA10 | 15 | 5 | 75 | 60 | 525 | 32.08 | 0.00% | 32.68 | 0.00% | 0.00% | −1.87% |
| LA11–LA15 | 20 | 5 | 100 | 80 | 950 | 11.81 | 11.81% | 12.29 | 0.00% | 11.81% | −4.06% |
| LA16–LA20 | 10 | 10 | 100 | 90 | 450 | 791.3 | 4.51% | 804.68 | 0.39% | 4.12% | −1.69% |
| LA21–LA25 | 15 | 10 | 150 | 135 | 1050 | 1158.66 | 10.35% | 1213.15 | 1.41% | 8.94% | −4.70% |
| LA26–LA30 | 20 | 10 | 200 | 180 | 1900 | 1595.35 | 9.46% | 2131.7 | 2.32% | 7.14% | −33.62% |
| LA31–LA35 | 30 | 10 | 300 | 270 | 4350 | 205.4 | 34.41% | 219.9 | 5.86% | 28.55% | −7.06% |
| LA36–LA40 | 15 | 15 | 225 | 210 | 1575 | 2961.47 | 15.38% | 3932.53 | 4.35% | 11.03% | −32.79% |
| ABZ5–ABZ6 | 10 | 10 | 100 | 90 | 450 | 700.91 | 13.67% | 747.51 | 4.41% | 9.26% | −6.65% |
| ABZ7–ABZ9 | 20 | 15 | 300 | 280 | 2850 | 1458.07 | 16.15% | 1552.67 | 6.28% | 9.87% | −6.49% |
| ORB01–ORB10 | 10 | 10 | 100 | 90 | 450 | 137.88 | 1.67% | 145.54 | 0.98% | 0.69% | −5.56% |

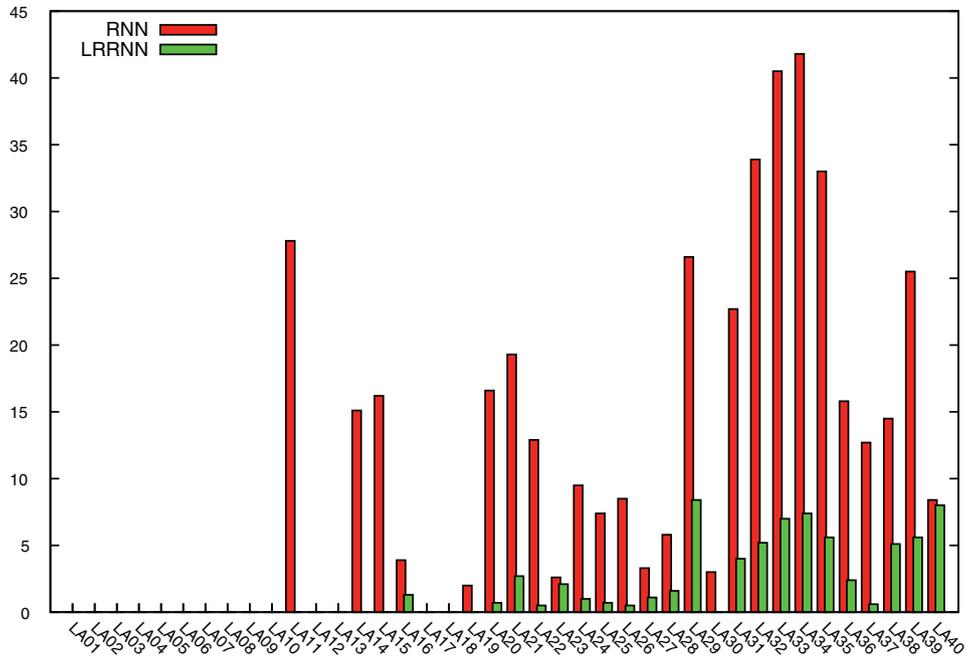

**Fig. 5.** Percentage deviation from optimal solution for LA problems.

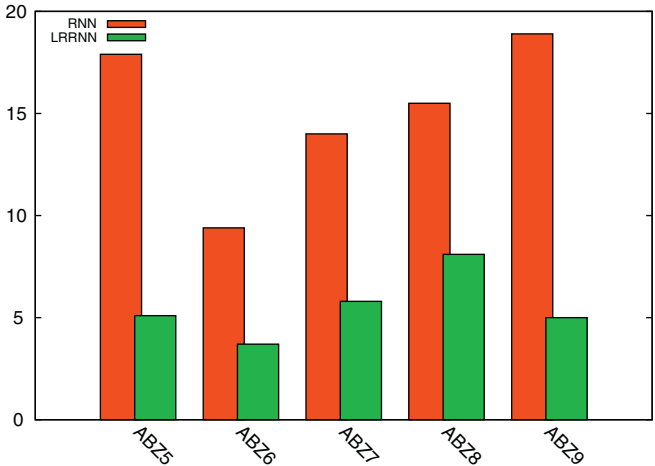

**Fig. 6.** Percentage deviation from optimal solution for ABZ problems.

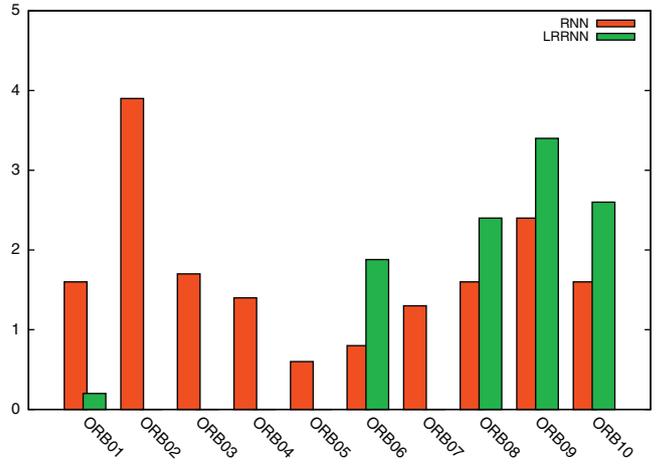

**Fig. 7.** Percentage deviation from optimal solution for ORB problems.

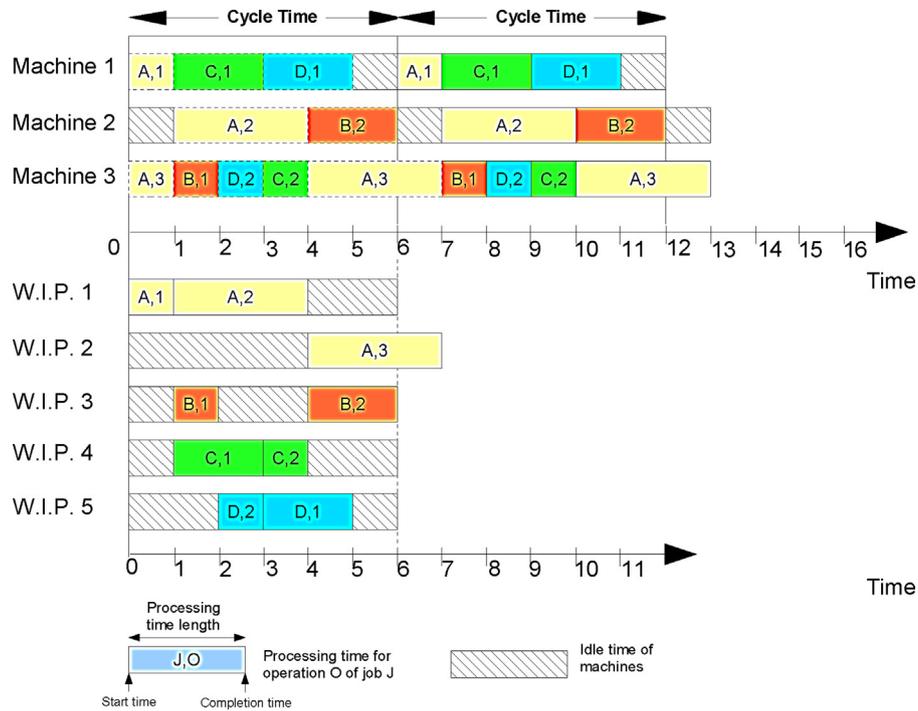

**Fig. 8.** Solution for CFMS problem HIL88 [17] from Advanced Hopfield Approach.

**Table 7**
Solution for FMS test problems with minimum WIP.

| FMS benchmarks | | | | | | | | | |
|---|---|---|---|---|---|---|---|---|---|
| Problem | N | M | O | CT | Feasible solutions | $WIP_{LB}$ | $WIP_{best}$ RNN | $WIP_{best}$ LRRNN | $WIP_{best}$ AdvHopfield |
| HIL87 | 4 | 4 | 15 | 8 | 1.6E+7 | 5 | **5** | **5** | **5** |
| HIL88 | 4 | 3 | 9 | 6 | 1.1E+4 | 5 | **5** | **5** | **5** |
| VAL94 | 3 | 5 | 13 | 11 | 7.9E+7 | 5 | 6 | **5** | **5** |
| OHL95 | 2 | 6 | 10 | 28 | 3.5E+10 | 4 | 5 | 5 | 5 |
| KORBAA98a | 7 | 9 | 23 | 24 | 8.6E+10 | 12 | 14 | 14 | 14 |
| KORBAA98b | 2 | 9 | 23 | 24 | 8.6E+10 | 9 | 12 | 10 | **9** |
| FT63(FT06) | 6 | 6 | 36 | 43 | 4.4E+38 | 7 | 9 | **7** | **7** |
| LA84(LA04) | 10 | 5 | 50 | 537 | 3.9E+71 | 10 | 12 | 12 | **10** |

**Table 8**
Comparison of solutions for FMS test problems.

| FMS benchmarks | | | | | | | | | | | |
|---|---|---|---|---|---|---|---|---|---|---|---|
| Problem | $WIP_{LB}$ | GA $WIP_{best}$ | GA $CPU_{best}$ | RNN $WIP_{best}$ | RNN $Iteration_{Best}$ | RNN $CPU_{best}$ | LRRNN $WIP_{best}$ | LRRNN $Iteration_{Best}$ | LRRNN $CPU_{best}$ | Advanced Hopfield $WIP_{best}$ | Advanced Hopfield $Iteration_{Best}$ | Advanced Hopfield $CPU_{best}$ |
| HIL87 | 5 | 5 | ~1 | **5** | 4530 | 13.9 | **5** | 9654 | 31.8 | **5** | 9020 | 16.6 |
| HIL88 | 5 | 5 | ~1 | **5** | 3490 | 10.7 | **5** | 7340 | 24.2 | **5** | 6530 | 12.3 |
| VAL94 | 5 | 5 | ~1 | 6 | 13,420 | 41.2 | **5** | 31,069 | 102.3 | **5** | 9870 | 18.2 |
| OHL95 | 4 | 5 | ~1 | 5 | 1,320,155 | 4056.0 | 5 | 288,716 | 950.4 | 5 | 29,098 | 53.5 |
| KORBAA98a | 12 | 13 | ~10 | 14 | 1,327,966 | 4080.0 | 14 | 391,090 | 1287.4 | 14 | 366,848 | 675.0 |
| KORBAA98b | 9 | 9 | ~60 | 12 | 1,230,322 | 3780.0 | 10 | 877,174 | 2887.5 | **9** | 1,020,652 | 1878.0 |
| FT63 (FT06) | 7 | 8 | ~48 h | 9 | 45,111,803 | 138,600.0 | **7** | 37,183,062 | 122,400.0 | **7** | 44,152,174 | 81,240.0 |
| LA84 (LA04) | 10 | 12 | ~72 h | 12 | 60,930,228 | 187,200.0 | 12 | 56,300,635 | 185,331.6 | **10** | 94,558,000 | 173,880.0 |

## 5. Conclusion

In this paper we studied the cyclic scheduling problem using neural network models. We propose the recurrent neural network (RNN) approach. This RNN approach tries to find optimum solutions by minimising the energy state of the network. We have derived the equations of motion for the network and developed some algorithms to deal with special situations. We have also extended this RNN technique by coupling it with the Lagrangian Relaxation method. The resulting approach is called Lagrangian Relaxation Recurrent Neural Network (LRRNN). This new approach obviously combines the abilities of the recurrent neural network and the Lagrange multipliers. One of the major characteristics of LRRNN is that it relaxes the constraints of the cyclic scheduling problem and, therefore, reduces the complexity of the problem.

The cyclic job shop scheduling is common in nearly every manufacturing processes. Our solution has reduced the gap between the theoretical studies and practical problem that we can encounter in real-world processes. For instance, the theoretical problem was always modelled and then simplified in order to relax its complexity and becomes easier to solve using traditional optimisation techniques. However in practice the complexity of the process remains the same and the adoption of solutions of the simplified problem are far from being of practical use. Our approach can cope with some complex dimensions of the cyclic job scheduling, such as the problem size, some level of constraints' complexity, and get very good solutions.

Experimental results show that both approaches (RNN and LRRNN) are efficient and return very good solutions for cyclic job shop scheduling problems. We found out that RNN does not scale well with the size of the problem. It does reasonably well for the problems of size up to about 100 operations. In other words, RNN is much more efficient on smaller problems, while LRRNN can deal much better with larger problems. In fact LRRNN performed very well on very large problems with simple constraints. However, it is still struggling with problems where the constraints are very complex. Currently the penalty function $\varphi()$ is a simple quadratic function, which may not be able to deal with very large set of constraints and their characteristics. We will look at other forms of functions and see how can be improved. Another possible solution is to model the system as a multi-objective problem and each of the objectives will be assigned to a recurrent neural network. The overall outputs will be fed to either another RNN or LRRNN. These models will be studied in the near future.

An extension of this research would involve researching to solve other types of cyclic scheduling problems that may include robotics problems, hoist problems, cyclic flow shop problem or cyclic open shop problem. This will greatly enhance the application of these techniques further and improve their internal processes for moving from one solution to another.

As further future work, we would like to study the sensitivity of these models to stochasticity in cyclic scheduling problems. The modelling of machine related issues, such as machines with bottlenecks, maintenance related issues, non-uniform speed of machines, or load variation on production lines, would allow these models to be refined and move a step closer to modelling real-life cyclic scheduling problems.

## References

[1] Brucker P. Scheduling Algorithms. Secaucus, NJ, USA: Springer-Verlag New York, Inc.; 2001.
[2] Hanen C. Study of a NP-hard cyclic scheduling problem: The recurrent job-shop. European Journal of Operational Research 1994;72(1):82–101.
[3] Brucker P, Kampmeyer T. Tabu search algorithms for cyclic machine scheduling problems. Journal of Scheduling 2005;8(4):303–22.
[4] Roundy R. Cyclic schedules for job shops with identical jobs. Mathematics of Operational Research 1992;17(4):842–65.
[5] Draper D, Jónsson A, Clements D, Joslin D. Cyclic scheduling. In: Sixteenth International Joint Conference on Artificial Intelligence (IJCAI '99). 1999. p. 1016–21.
[6] Kampmeyer T. Cyclic scheduling problems. PhD thesis, University of Osnabruck, Germany; 2006.
[7] Hitz K. Scheduling of flexible flow shops. Tech. Rep., MIT, Cambridge, MA; 1979.
[8] Metaxiotis K, Psarras J. Neural networks in production scheduling: intelligent solutions and future promises. Applied Artificial Intelligence 2003;17(4):361–73.
[9] Sabuncuoglu I. Scheduling with neural networks: a review of the literature and new research directtions. Production Planning and Control 1998;9(1):2–12.
[10] Huang S, Zhang H-C. Neural networks in manufacturing: a survey. In: Fifteenth IEEE/CHMT international on electronic manufacturing technology symposium. 1993. p. 177–90, http://dx.doi.org/10.1109/IEMT.1993.398204.
[11] Cichocki A, Unbehauen R, Weinzierl K, Holzel R. A new neural network for solving linear programming problems. European Journal of Operational Research 1996;93(2):244–56.
[12] Fisher H, Thompson GL. Probabilistic learning combinations of local job-shop scheduling rules. In: Muth J, Thompson G, editors. Industrial scheduling. Englewood Cliffs: Prentice Hall; 1963. p. 225–51.
[13] Lawrence S. Resource constrained project scheduling: an experimental investigation of heuristic scheduling techniques (Supplement) Graduate School of Industrial Administration. Carnegie-Mellon University; 1984.
[14] Adams J, Balas E, Zawack D. The shifting bottleneck procedure for job shop scheduling. Management Science 1988;34(3):391–401.
[15] Applegate D, Cook W. A computational study of the job-shop scheduling problem. ORSA Journal on Computing 1991;3(2):149–56.
[16] Storer R, Wu S, Vaccari R. New search spaces for sequencing problems with application to job shop scheduling. Management Science 1992;38(10):1495–509.
[17] Hillion H, Proth J. Analyse de fabrications non linéaires et répétitives à l'aide des graphes d'evènements temporisès. RAIRO 22 (2).
[18] Panwalker SS, Iskander W. A Survey of Scheduling Rules. Journal of Operations Research 1977;25(1):45–61.
[19] Conway RW, Maxwell WL, Miller LW. Theory of Scheduling. Addison-Wesley; 1967.